%% file: DGRL-colt2020.tex
\documentclass{colt2020} 

\title{Dual Graph Representation Learning }
\usepackage{times}

\usepackage{latexsym}
\input{math_commands.tex}

\usepackage{hyperref}

\usepackage{times}
\usepackage{xcolor}
\usepackage{color}
\usepackage{soul}
\usepackage[utf8]{inputenc}
\usepackage[small]{caption}
\usepackage{algorithm}
\usepackage{algorithmic}
\usepackage{boxedminipage}
\usepackage{multirow} 
\usepackage{makecell}
\usepackage{algorithm,algorithmic}

\newcommand{\ignore}[1]{{}}
\usepackage{multirow}

\coltauthor{%
 \Name{Huiling Zhu} \Email{zhuhl02@gmail.com}\\
 \addr 
 \AND
 \Name{Xin Luo} \Email{luo35@mail2.sysu.edu.cn}\\
 \addr Sun Yat-Sen University, Guangzhou, China
 \AND
 \Name{Hankz Hankui Zhuo} \Email{zhuohank@mail.sysu.edu.cn}\\
 \addr Sun Yat-Sen University, Guangzhou, China
}

\begin{document}

\maketitle

\begin{abstract}%
Graph representation learning embeds nodes in large graphs as low-dimensional vectors and  is of great benefit to many downstream applications. Most embedding frameworks, however, are inherently transductive and unable to  generalize to unseen nodes or learn representations across different graphs. Although inductive approaches can generalize to unseen nodes, they neglect different contexts of nodes and cannot learn node embeddings dually. In this paper, we present a context-aware unsupervised dual encoding framework, \textbf{CADE},  to generate  representations of nodes by combining real-time neighborhoods  with neighbor-attentioned representation, and preserving extra memory of known nodes. We exhibit that our approach is effective by comparing to state-of-the-art methods.
\end{abstract}

\begin{keywords}%
graph representation learning, unsupervised learning, dual learning
\end{keywords}

\section{Introduction}

The study of real world graphs, such as social network analysis (\cite{DBLP:journals/debu/HamiltonYL17}), molecule screening (\cite{DBLP:conf/nips/DuvenaudMABHAA15}),  knowledge base reasoning (\cite{DBLP:conf/icml/TrivediDWS17}), and biological protein-protein networks analysis (\cite{DBLP:journals/bioinformatics/ZitnikL17}), evolves with the development of computing technologies. Learning  vector representations of graphs is effective for a  variety of prediction and graph analysis tasks (\cite{DBLP:conf/kdd/GroverL16,DBLP:conf/www/TangQWZYM15}). 
High-dimensional information about neighbors of nodes is represented by dense vectors, which can be fed to off-the-shelf approaches to solve tasks, such as node classification (\cite{DBLP:conf/aaai/WangCWP0Y17,DBLP:books/sp/social2011/BhagatCM11}), link prediction (\cite{DBLP:conf/kdd/PerozziAS14,DBLP:conf/www/WeiXCY17}), node clustering (\cite{DBLP:conf/aaai/NieZL17,DBLP:conf/icdm/DingHZGS01}), recommender systems (\cite{DBLP:conf/kdd/YingHCEHL18}) and visualization (\cite{maaten2008visualizing}).

There are mainly two types of models for graph representation learning. Transductive approaches (\cite{DBLP:conf/kdd/PerozziAS14,DBLP:conf/kdd/GroverL16,DBLP:conf/www/TangQWZYM15}) are able to learn representations of existing nodes but unable to generalize to new nodes. However, in real-world evolving graphs such as social networks, new users which join in the networks dynamically must be represented based on the representations of existing nodes. Inductive approaches were proposed to address this issue. 
GraphSAGE (\cite{hamilton2017inductive}),  a hierarchical sampling and aggregating framework, successfully leverages feature information to generate embeddings of  new nodes. However, it samples all neighborhood nodes randomly and uniformly without considering the difference of nodes. GAT (\cite{DBLP:journals/corr/abs-1710-10903}) uses given class labels to guide attention over neighborhoods so as to aggregate useful feature information. 
However, without  ground-truth class labels, it is difficult for unsupervised approaches to build attention. 

\begin{figure*}[!th]
	\centering
	\includegraphics[width=0.65\textwidth]{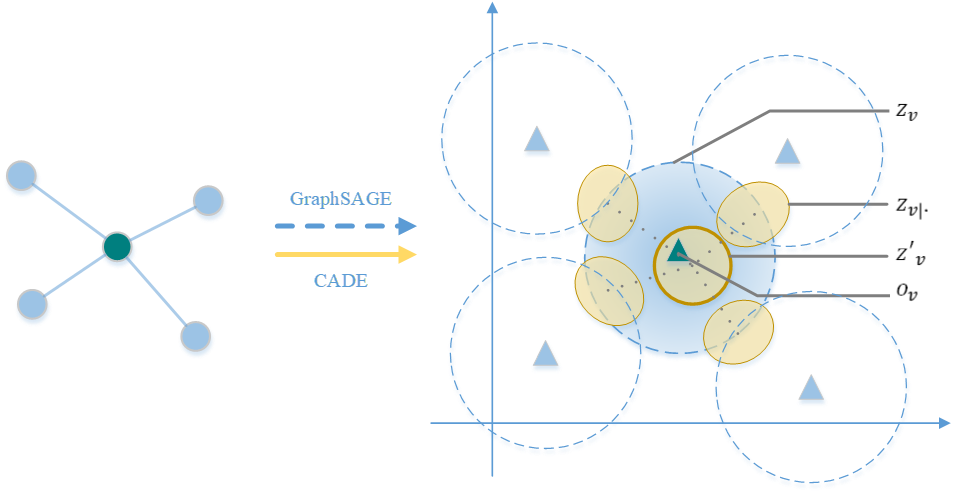}
	\caption{Visual comparison between representations learnt by current methods and dual encoding.}
	\label{pe1}
\end{figure*}

In this paper, we introduce a \textbf{dual encoding} framework for unsupervised inductive representation learning of graphs. Instead of learning self-attention over neighborhoods of nodes, we exploit  bi-attention between representations of two nodes that co-occur in a short random-walk (\textbf{positive pair}). Figure \ref{pe1} illustrate the embedding of nodes into low-dimensional vectors, where  each node $v$ has an optimal embeddings $o_v$. Yet the direct output of encoder $z_v$ of GraphSAGE could be located anywhere.
Specifically, given feature input from both sides of a positive pair $(v, v_p)$,  a neural network is trained to encode the pair into $K$ different embeddings $z^k_v$ and $z^k_{v_p}$ through different sampled neighborhoods or different encoding functions. Then, a bi-attention layer is applied to generate the most adjacent matches $z_{v|v_p}$ and $z_{v_p|v}$, which will be referred as dual-representations. By putting most attention on  the pair of embeddings with smallest difference,  dual representation of nodes with less deviation will be generated, which can be visualized as $z_{v|\cdot}$ in Figure \ref{pe1}.

GraphSAGE  assumes that unseen nodes can be (easily) represented by known graphs data.
We combine the ground truth structure and the learned dual-encoder to generate final representation. Unseen nodes can be represented based on their neighborhood structure. 
Current inductive approaches have no direct memory of the training nodes. We combine the idea of both transductive and inductive approaches via associating an additive global embedding bias to each node, which can be seen as a memorable global identification of each node in training sets.


Our contributions include: (1)we introduce a dual encoding framework to produce context-aware representation for nodes, and conduct experiments to demonstrate its efficiency and effectiveness, (2)we apply bi-attention mechanism for graph representation dual learning, managing to learn dual representation of nodes more precisely and (3)we combine the training of transductive global bias with inductive encoding process, as memory of nodes that are already used for training.

\section{Related Work}
Following (\cite{DBLP:journals/tkde/CaiZC18}, \cite{DBLP:journals/corr/abs-1906-02989} and \cite{DBLP:journals/kbs/GoyalF18}), there are mainly two types of approaches:
\subsection{Network embedding}
For unsupervised embedding learning, DeepWalk (\cite{DBLP:conf/kdd/PerozziAS14}) and node2vec (\cite{DBLP:conf/kdd/GroverL16}) are based on random-walks extending the Skip-Gram model; LINE (\cite{DBLP:conf/www/TangQWZYM15})seeks to preserve first- and second-order proximity and trains the embedding via negative sampling; SDNE (\cite{DBLP:conf/kdd/WangC016}) jointly uses unsupervised components to preserve second-order proximity and expolit first-order proximity in its supervised components; TRIDNR (\cite{DBLP:conf/ijcai/PanWZZW16}), CENE(\cite{DBLP:journals/corr/SunGDL16}), TADW (\cite{DBLP:conf/ijcai/YangLZSC15}),GraphSAGE (\cite{hamilton2017inductive}) utilize node attributes and potentially node labels.
Convolutional neural networks are also applied to  graph-structured data. For instance, GCN (\cite{DBLP:conf/iclr/KipfW17}) proposed an simplified graph convolutional network. These graph convolutional network based approaches are (semi-)supervised. Recently, inductive graph embedding learning (\cite{hamilton2017inductive} \cite{DBLP:journals/corr/abs-1710-10903} \cite{DBLP:journals/corr/BojchevskiG17} \cite{DBLP:conf/icdm/Derr0T18} \cite{DBLP:conf/kdd/GaoWJ18}, \cite{DBLP:conf/aaai/LiHW18}, \cite{DBLP:conf/aaai/WangWWZZZXG18} and \cite{DBLP:conf/nips/YingY0RHL18}) produce impressive performance across several large-scale benchmarks. 

\subsection{Attention}

Attention mechanism in neural processes have been extensively studied in neuroscience and computational neuroscience (\cite{itti1998model,desimone1995neural}) and frequently applied in deep learning for speech recognition (\cite{chorowski2015attention}), translation (\cite{luong2015effective}), question answering (\cite{DBLP:journals/corr/SeoKFH16}) and visual identification of objects (\cite{DBLP:conf/icml/XuBKCCSZB15}).  Inspired by (\cite{DBLP:journals/corr/SeoKFH16} and \cite{DBLP:conf/nips/Abu-El-HaijaPAA18}), we construct a bi-attention layer upon aggregators to capture useful parts of the neighborhood.

\section{Model}

Let $\gG=\{V,E,\mX\}$ be an undirected graph, where a set of nodes $V$ are connected by a set of edges $E$, and $\mX\in \R^{|V|\times f}$ is the attribute matrix of nodes. A global embedding bias matrix is denoted by $\mB \in R^{|V|\times d}$, where a row of $\mB$ represents the d-dimensional global embedding bias of a node. The hierarchical layer number, the embedding output of the $l$-th layer and the final output embedding are denoted by $L$, $\vh^l$ and $\vz$, respectively.

\subsection{Context-aware inductive embedding encoding}\label{ca}

The embedding generation process is described in Algorithm \ref{cade}. Assume that the dual encoder is trained and parameters are fixed. After training, positive pairs are collected by random walks on the whole dataset. The features of each positive pair are passed through a dual-encoder. Embeddings of nodes are generated so that the components of a pair are related and adjacent to each other.

\begin{algorithm}[!ht]	
	\caption{Context-Aware Dual-Encoding (\textbf{CADE})}\label{cade}
	\textbf{input:} the whole graph $\gG=(V,E)$; the feature matrix $\mX$; the trained $\rm{DualENC}$\\
	\textbf{output:} learned embeddings $\vz$;
	\begin{algorithmic}[1]
		\STATE Run random walks on $\gG$ to gain a set of positive pair $\sP$;
		\STATE $\sZ_v \leftarrow \emptyset, \forall v \in V$ 
		\FOR{$(v,v_p) \in \sP$}
		\STATE $z_v, z_{v_p} = \rm{DualENC}(v, v_p, \gG, \mX)$;
		\STATE $\sZ_v \leftarrow \sZ_v \cup z_v$
		\STATE $\sZ_{v_p} \leftarrow \sZ_{v_p} \cup z_{v_p}$
		\ENDFOR
		\FOR{$v \in V$}
		\STATE $\vz_v = \rm{Mean}(\sZ_v)$;
		\ENDFOR
	\end{algorithmic}
\end{algorithm}

\subsection{Dual-encoder with multi-sampling}\label{dems}
\begin{figure}[!ht]
	\centering
	\includegraphics[width=0.45\textwidth]{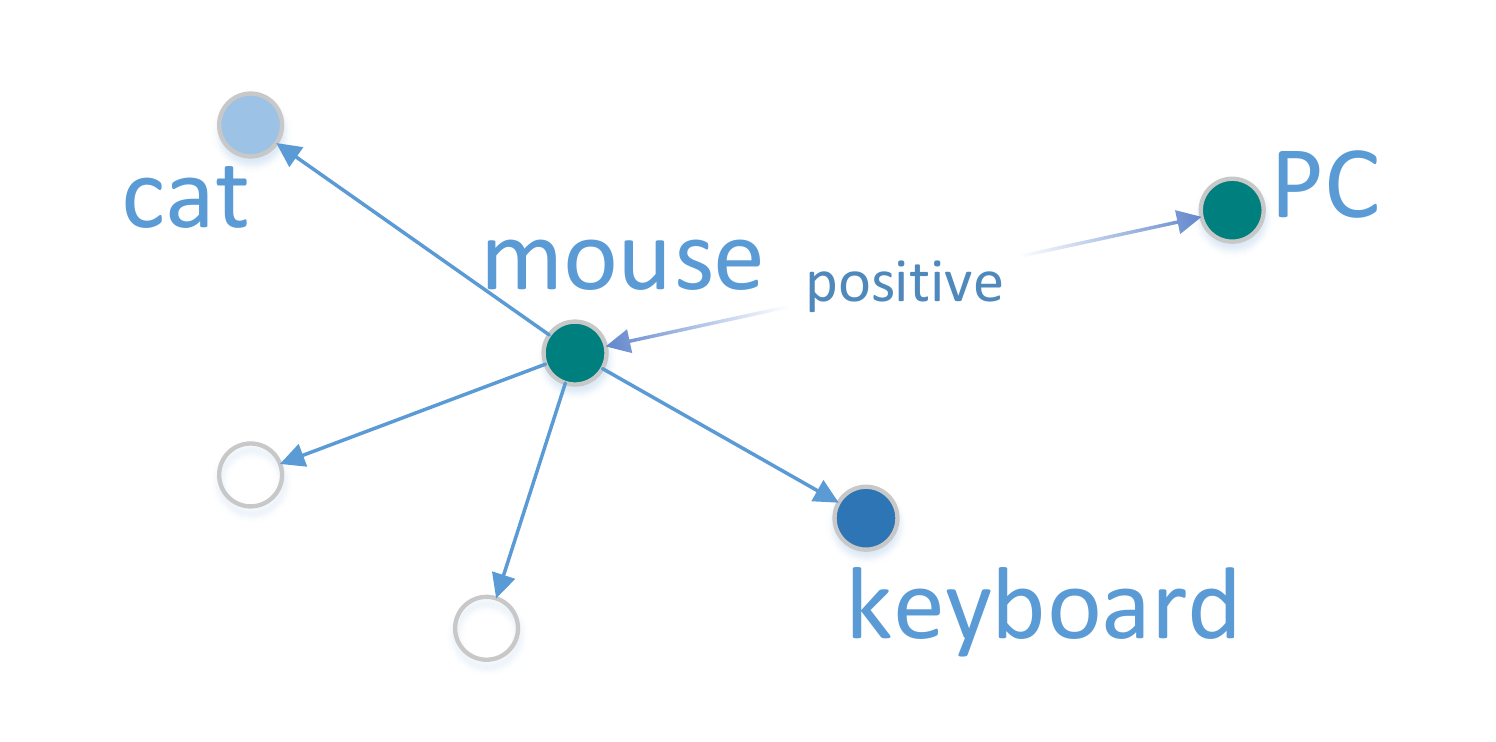}
	\caption{Visualization of the choice of neighborhood/perspective for encoding embedding.}
	\label{example}
\end{figure}

In this subsection, we explain the dual encoder. In the hierarchical sampling and aggregating framework (\cite{hamilton2017inductive}), it is challenging and vital to select relevant neighbor with the layer goes deeper and deeper.  
For example, as shown in Figure \ref{example}, given word "mouse" and its positive node "PC", it is better to sample "keyboard", instead of "cat", as a neighbor node. However, to sample the satisfying node according to heuristic rules layer by layer is very time consuming, and it is difficult to learn attention over neighborhood for unsupervised embedding learning.

As a matter of fact, these neighbor nodes are considered to be useful because they are more welcome to be sampled as input so as to produce more relevant output of the dual-encoder. 
Therefore, instead of physically sampling these neighbor nodes, in Step 2 to Step 6 in Algorithm \ref{de}, we directly apply a \textbf{bi-attention layer} on the two sets of embedding outputs with different sampled neighborhood feature as input, so as to locate the most relevant representation match, as a more efficient approach to exploring the most useful neighborhood.

\begin{algorithm}[!ht]	
	\caption{DualENC}\label{de}
	\textbf{input:} Training graph $\gG(V,E)$; node attributes $\mX$; global embedding bias matrix $\mB$; sampling times $K$; positive node pair $(v, v_p)$;\\
	\textbf{output:} adjacent embeddings $z_v$ and $z_{v_p}$;
	\begin{algorithmic}[1]
		\STATE For node $v$ and $v_p$, generate $K$ embeddings, $\vh_v$, $\vh_{v_p}$, using a base encoder $\rm{SAGB}$
		\FOR{$i,j \in \{1,...,K\}$}
		\STATE $S_{i,j} \leftarrow \alpha (\vh_{vi},\vh_{v_pj})$
		\ENDFOR
		\STATE softmax on flattened similarity matrix $S$: $S_{i,j} \leftarrow \frac{e^{S_{i,j}}}{\sum_{0,0}^{K,K}e^{S_{i,j}}}$
		\STATE calculate attention $a_v$ and $a_{v_p}$: $\bm{a}_{vi} \leftarrow \sum_{j=1}^{K}S_{i,j},\bm{a}_{v_pj} \leftarrow \sum_{i=1}^{K}S_{i,j}$
		\STATE $\vz_{v} \leftarrow \sum_{t=1}^K \bm{a}_{vk}\vh^{L}_{vk}$
		\STATE $\vz_{v_p} \leftarrow \sum_{t=1}^K \bm{a}_{v_pk}\vh^{L}_{v_pk}$
	\end{algorithmic}
\end{algorithm}

\begin{figure*}[!ht]
	\centering
	\includegraphics[width=0.9\textwidth]{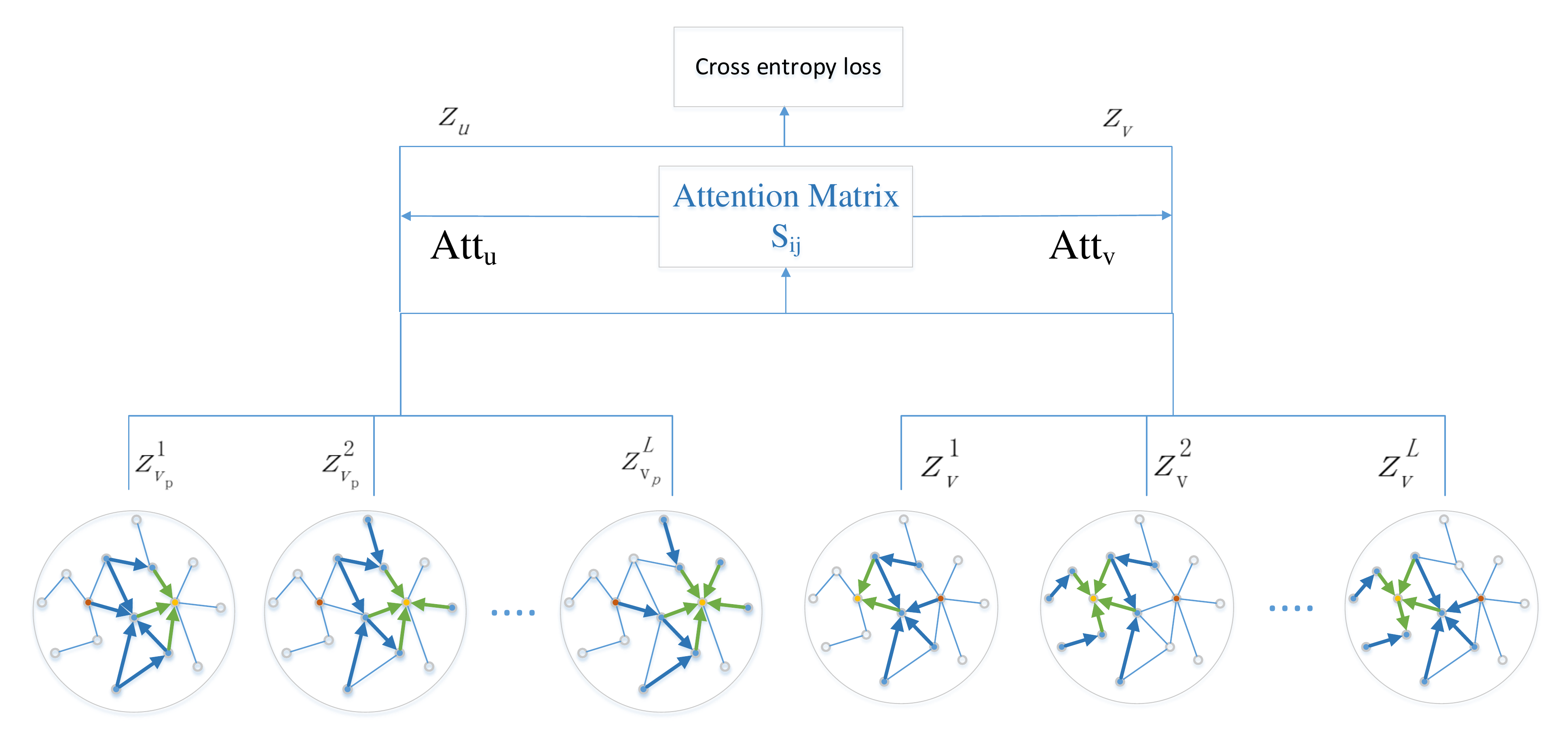}
	\caption{Bi-attention layer between the final aggregating layer and loss layer.}
	\label{biattention}
\end{figure*}

We use the hierarchical sampling and aggregating framework as a base encoder in our experiments, but it can also be designed in many other ways. The aggregation process with bi-attention architecture is illustrated by Figure \ref{biattention}.

To train our encoder before using it to generate final representations of nodes, we apply a typical pair reconstruction loss function with negative sampling \cite{hamilton2017inductive}: 
\begin{equation}
	J_{\gG}(z_v) = -\mathrm{log}(\sigma(\vz_v^T\vz_{v_p})) - Q\cdot \E_{v_n\sim P_n(v)}\mathrm{log}(\sigma(-\vz_v^T\vz_{v_n}))
\end{equation}
where node $v_p$ co-occurs with $v$ on fixed-length random walk (\cite{DBLP:conf/kdd/PerozziAS14}), $\sigma$ is the sigmoid function, $P_n$ is a negative sampling distribution, $Q$ defines the number of negative samples. Note that $z_v$ and $ z_{v_p}$ are dual representation to each other while $z_{v_n}$ represents the direct encoder output of negative sample $v_n$.

\subsection{Dual-encoder with multi-aggregating}\label{dema}
Besides learning dual representation with multiple sampling, we introduce another version of our dual encoder with multiple aggregator function.The intuition is that through different perspective, a node can be represented differently corresponding to different kinds of positive nodes. For example, when encoding "mouse" for positive node "PC", the ideal aggregator is to focus on encoding features about digital products instead of animals. 

In Step 1 in Algorithm \ref{de}, for a node $v$, we sample neighborhood once and aggregate feature with $K$ sets of parameters, gaining $K$ different representations $h_{vk}$ corresponding to $K$ different character of $v$. Given a positive node pair, $v$ and $v_p$, their dual representation are calculated by applying bi-attention as we described in the last section. There is but one difference that we use a weigh vector $\bm{A} \in R^{2d}$ as parameter instead of dot-product, to calculate the $K\times K$ attention matrix between node $v$ and node $v_p$: $
	S_{ij} \leftarrow \frac{\mathrm{exp}(\bm{A}^\mathsf{T}[\vh_{vi}||\vh_{v_pj}])}{\sum_{0,0}^{K,K}\mathrm{exp}(\bm{A}^\mathsf{T}[\vh_{vi}||\vh_{v_pj}])}
$, where $\cdot^T$ represents transposition and $||$ is the concatenation operation. 
The rest of calculation of dual representation is same as section \ref{dems}.

\begin{figure}[!ht]
	\centering
	\includegraphics[width=0.5\textwidth]{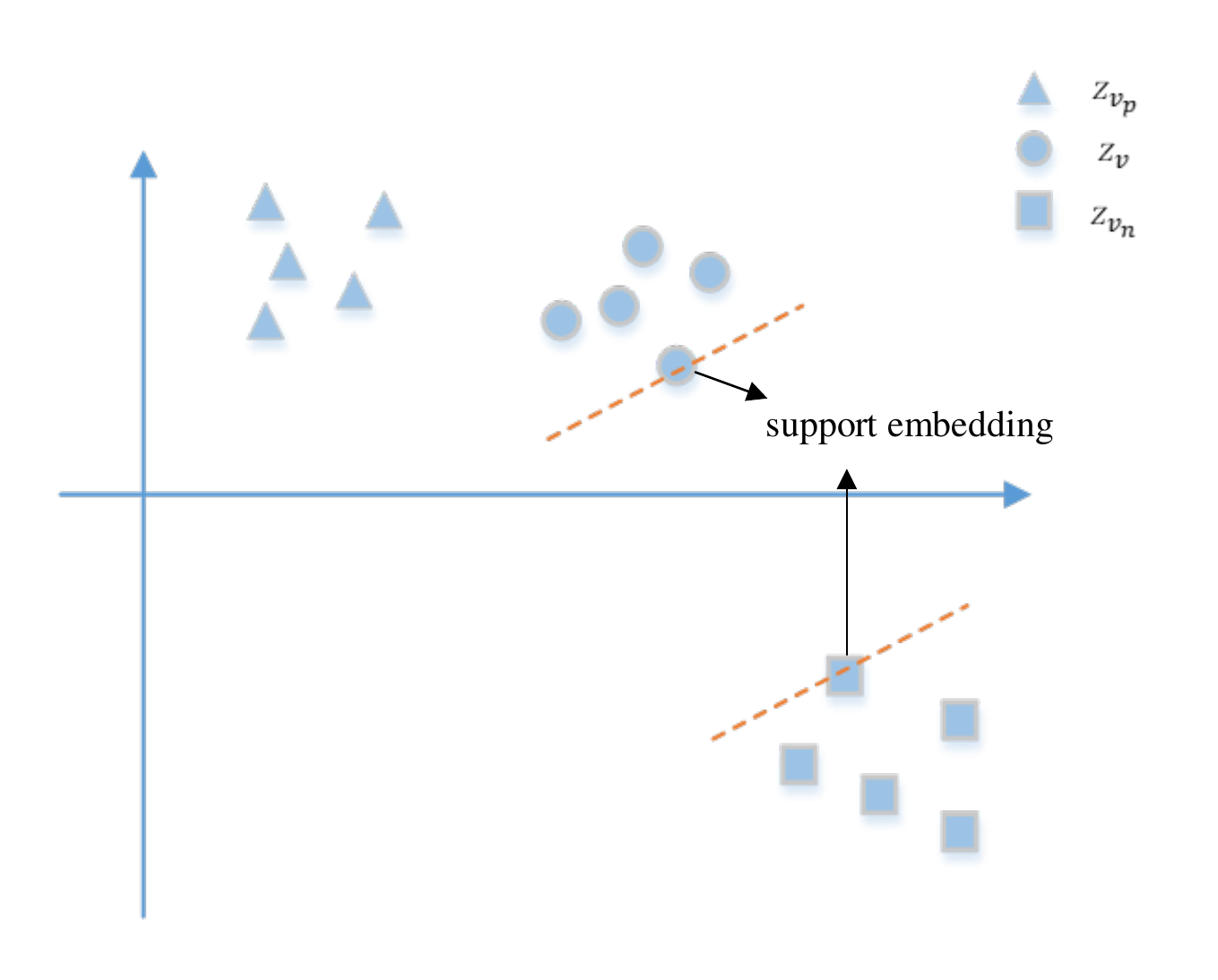}
	\caption{Training support embeddings of node $v$ and its negative sample node $v_n$.}
	\label{dema-train}
\end{figure}

Another difference is during training. 
With $K$ sets of parameter for aggregating, negative sample $v_n$ is now also represented by $K$ different embeddings. As shown in Figure \ref{dema-train}, we set $K=5$ and use different shape to represent the embeddings of the positive node pair and the negative sampled nodes.

As we can see in Figure \ref{dema-train}, to make sure that any embeddings of node $v_n$ as far away from any of node $v$ as possible, it is equal to maximizing the distance between their \textbf{support embeddings}, which is the closest pair of embeddings of $v$ and $v_n$. The support embedding can be calculated by the learned dual encoder. In conclusion, our loss function can be modified as follows:

\begin{align}
	J_{\gG}(z_v) &= -\mathrm{log}(\sigma(\vz_v^T\vz_{v_p})) - Q\cdot \E_{v_n\sim P_n(v)}\mathrm{log}(\sigma(-\vz^{'\mathsf{T}}_v\vz_{v_n}))\\
	\vz_v, \vz_{v_p} &= \mathrm{DualENC}(v, v_p, \bm{A})\\
	\vz'_v, \vz_{v_n} &= \mathrm{DualENC}(v, v_p, \bm{A}^*)
\end{align}
where $\bm{A}^*$ representing that we stop the back-propagation through $\bm{A}$ in dual encoding for negative sample node, since $\bm{A}$ are supposed to learn bi-attention between the positive node pair and be reused only to capture the support embedding of $v$ and its negative sample nodes.

\subsection{Memorable global bias in hierarchical encoding}\label{gb}

In this section, we first explain the base encoder used in our proposed dual encoding framework, and then we introduce how we apply memorable global bias within this framework.

The general intuition of GraphSAGE is that at each iteration, nodes aggregate information from their local neighbors, and as this process iterates, nodes incrementally gather more and more information from further reaches of the graph.
For generating embedding for one specific node $u$, we describe the process below.
First, we construct a neighborhood tree with node $u$ as the root, $\sN_u$, by iteratively sampling immediate neighborhood of nodes of the last layer as children. Nodes at the $l$th layer are represented by symbol $\sN_u^l$, $\sN_u^0 = \{u\}$.
Then, at each iteration, each node $i$ aggregates the representations of its children $j$, $\{h_j^{l-1}\}$, and of itself, $h_i^{l-1}$, into a single vector $h_i^l$, as representation of the next layer.
After $L$ iterations, we gain the $L$th layer representation of $v$, as the final output.

While this framework generates good representation for nodes, it cannot preserve sufficient embedding informations for known nodes. More specifically, for nodes that are known but trained less than average, the learned model would have treated them like nodes unmet before. Therefore, we intuitively apply distinctive and trainable global bias to each node, as follows:

\begin{align}
	\vh^{l-1}_{\gS(i)} &\leftarrow \mathrm{AGGREGATE}_l(\{\vh^{l-1}_j, \forall j\in \gS(i)\})\\
	\vh_i^l &\leftarrow \sigma(\mW^l \cdot [\vh_i^{l-1}||\vh^{l-1}_{\gS(i)}])\\
	\vh^{l}_i &\leftarrow \vh_i^l+\vb_i, l<L\\
	\vb_i &\leftarrow one\_hot(i)^\mathsf{T}\mB
\end{align}
where $\mB\in R^{|V|\times d}$ is the trainable global bias matrix, $\gS(i)$ represents the sampled neighborhood and also the children nodes of node $i$ in the neighborhood tree, AGGREGATE represents the neighborhood aggregator function, and $||$ is a operator of concatenating vectors.

On one hand, $\mB$ can be reused to produce embeddings for the known nodes or the unknown connected with the known, as supplement to the neural network encoder. On another hand, the global bias vectors can partially offset the uncertainty of the encoding brought by the random sampling not only during the training but also the final generation. Lastly but not least, we use only one set of global bias for all nodes, which means for any node, its representations of hidden layers are all added by the same bias vector. As a result of that, we are able to update parameters of aggregator function in the lowest layer with the global updated bias of nodes, highly increasing the training efficiency.

It is important for us to apply no global bias to the last layer of output, which is also the candidate of the dual-encoder output of nodes before applied with attention. The reason is that applying extra bias onto the last layer would directly change the embedding distribution of known nodes, making it unequal to the embedding distribution of unseen nodes.
\ignore{
	\begin{figure}[!ht]
		\centering
		\includegraphics[width=0.45\textwidth]{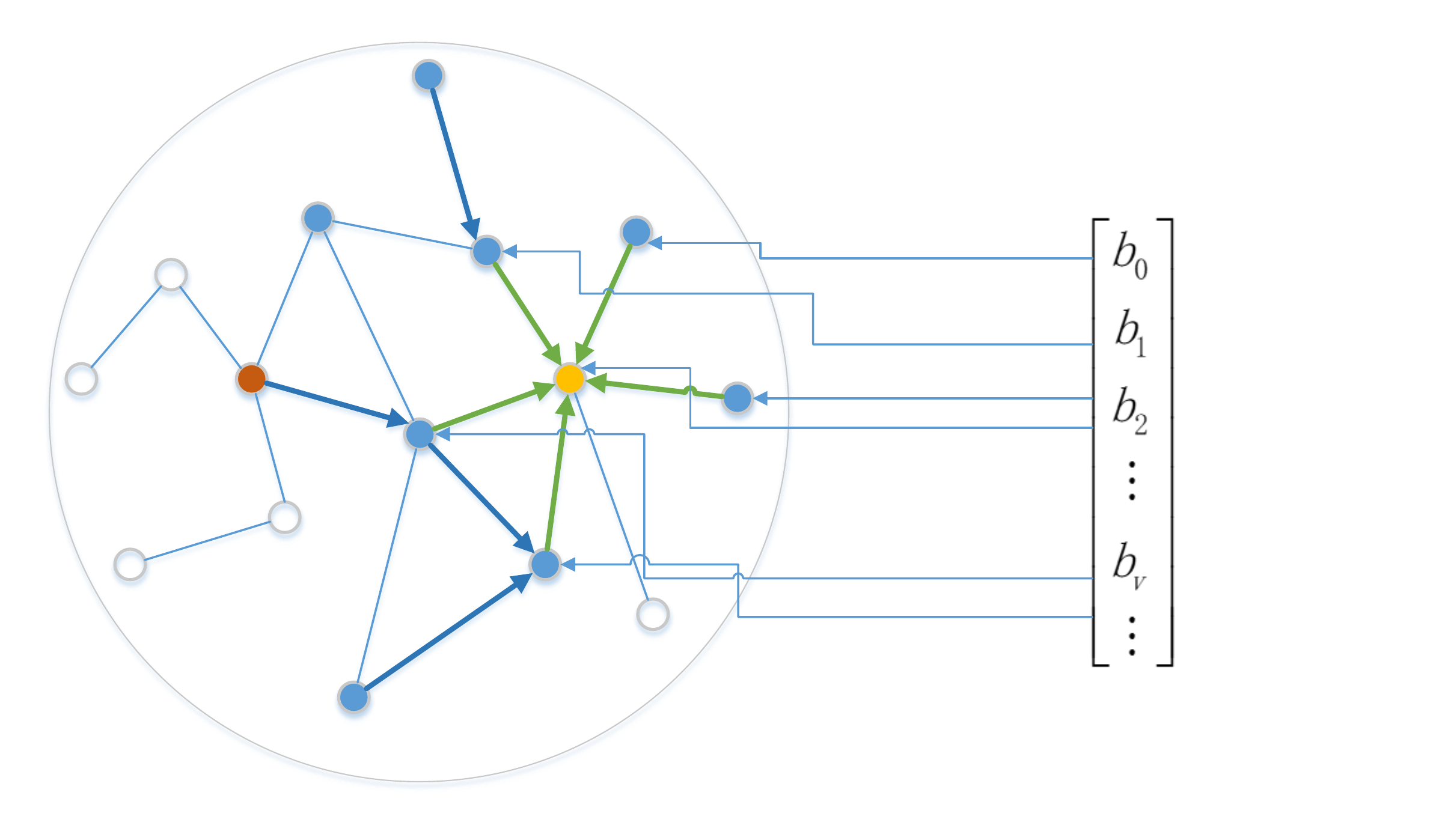}
		\caption{Applying global bias through aggregating .}
		\label{global-bias}
	\end{figure}
}
In general, the implementation of the base encoder with global bias is shown in Algorithm \ref{sagb}. The * in Step 8 means the children of node $i$ in the neighborhood tree $\sN_u$.

\begin{algorithm}[!ht]	
	\caption{SAGB:sampling and aggregating with global bias}\label{sagb}
	\textbf{input:} node $u$; hierarchical depth $L$; weight matrices $\mW^l$; non-linearity $\sigma$; differentiable neighbor aggregator $\mathrm{AGGREGATE}_l$; fixed-size uniform sampler $\gS: v \rightarrow 2^V$\\
	\textbf{output:} embedding $\vz_u$;
	\begin{algorithmic}[1]
		\STATE $\sN_u^0=\{u\}$;
		\FOR{$l=1...L$}
		\STATE $\sN_u^l\leftarrow \{\gS(i), \ \forall i \in \sN_u^{l-1}\}$;
		\ENDFOR
		\FOR{l=1...L}
		\FOR{$i \in \sN_u^0 \bigcup \sN_u^1 \bigcup ...\bigcup \sN_u^{L-l}$}
		\STATE $\vh^{l-1}_{\gS^*(i)} \leftarrow \mathrm{AGGREGATE}_l(\{\vh^{l-1}_j, \forall j\in \gS^*(i)\})$
		\STATE $\vh_i^l \leftarrow \sigma(\mW^l \cdot [\vh_i^{l-1}||\vh^{l-1}_{\gS^*(i)}])$
		\STATE \textbf{if} $l<L$: $\vh_i^l \leftarrow \vh_i^l + \mathrm{one\_hot}(i)^\mathsf{T}\mB $
		\ENDFOR
		\ENDFOR
		\STATE \textbf{return} $\vz_{u} \leftarrow \vh^L_{u}$
	\end{algorithmic}
\end{algorithm}

\section{Experiments}

In this section, we compare CADE against two strong baselines in an inductive and unsupervised setting, on challenging benchmark tasks of node classification and link prediction. We also perform further studies of the proposed model in section \ref{modelstudy}.

\subsection{Datasets}

The following graph datasets are used in experiments and statistics are summarized in Table\ref{ds}:
\begin{itemize}
	\item \textbf{Pubmed}: The PubMed Diabetes (\cite{DBLP:journals/aim/SenNBGGE08})\footnotemark[1] dataset is a citation dataset which consists of scientific publications from Pubemd database pertaining to diabetes classified into one of three classes.  Each publication in the dataset is described by a TF/IDF (\cite{DBLP:conf/sigir/SaltonY73}) weighted word vector from a dictionary. 
	\item \textbf{Blogcatalog}: BlogCatalog\footnotemark[2] is a social blog directory which manages bloggers and their blogs, where bloggers following each others forms the network dataset. 
	\item \textbf{Reddit}: Reddit\footnotemark[3] is an internet forum where users can post or comment on any content. We use the exact dataset conducted by (\cite{hamilton2017inductive}), where each link connects two posts if the same user comments on both of them. 
	\item \textbf{PPI}: The protein-protein-interaction (PPI) networks dataset contains 24 graphs corresponding to different human tissues(\cite{DBLP:journals/bioinformatics/ZitnikL17}). We use the preprocessed data also provided by (\cite{hamilton2017inductive}). 
\end{itemize}
\begin{table*}[!th]
	\centering
	\caption{Dataset Statistics}\label{ds}\scalebox{0.8}{
		\begin{tabular}{cccccc}
			\hline
			Dataset     & Nodes   & Edges      & Classes & Features & Avg Degree \\\hline
			Pubmed      & 19717   & 44324      & 3       & 500      & 4.47\\
			Blogcatalog & 5196    & 171743     & 6       & 8189     & 66.11\\
			Reddit      & 232,965 & 11,606,919 & 41      & 602      & 100.30\\
			PPI         & 56944   & 818716     & 121\footnotemark[4] & 50 & 28.76\\\hline
	\end{tabular}}
\end{table*}

\subsection{Experimental settings}
We compare CADE against the following approaches in a fully unsupervised and inductive setting:
\begin{itemize}
	\item GraphSAGE: In our proposed model, CADE, the base encoder mainly originates from GraphSAGE, a hierarchical neighbor sampling and aggregating encoder for inductive learning. Three alternative aggregators are used in Graphsage and CADE: (1) Mean aggregator, which simply takes the elementwise mean of the vectors in ${h^{k-1}_{u\in N(v)}}$; (2) LSTM aggregator, which adapts LSTMs to encode a random permutation of a node's neighbors' $h^{k-1}$; (3) Maxpool aggregator, which apply an elementwise maxpooling operation to aggregate information across the neighbor nodes.
	
	\item Graph2Gauss (\cite{DBLP:journals/corr/BojchevskiG17}): Unlike GraphSAGE and my method, G2G only uses the attributes of nodes to learn their representations, with no need for link information. Here we compare against G2G to prove that certain trade-off between sampling granularity control and embedding effectiveness does exists in inductive learning scenario.
\end{itemize}
Beside the above two models, we also include experiment results of raw features as baselines.
In comparison, we call the version of dual-encoder with multiple sampling as \textbf{CADE-MS}, while the version with multiple aggregator function as \textbf{CADE-MA}.

\footnotetext[1]{Available at https://linqs.soe.ucsc.edu/data.}
\footnotetext[2]{http://www.blogcatalog.com/}
\footnotetext[3]{http://www.reddit.com/}
\footnotetext[4]{PPI is a multi-label dataset.}

For CADE-MS, CADE-MA and GraphSAGE, we set the depth of hierarchical aggregating as $L=2$, the neighbor sampling sizes as $s_1=20, s_2=10$, and the number of random-walks for each node as 100 and the walk length as 4. 
The sampling time in CADE-MS or the number of aggregator in CADE-MA is set as $K=10$. 
And for all emedding learning models, the dimension of embeddings is set to 256, as for raw feature, we use all the dimensions. 
Our approach is impemented in Tensorflow (\cite{DBLP:journals/corr/AbadiABBCCCDDDG16}) and trained with the Adam optimizer (\cite{DBLP:journals/corr/KingmaB14}) at an initial learning rate of 0.0001. 

\begin{table*}[!th]
	\centering
	\caption{Prediction results for Pubmed/Blogcatalog w.r.t different unseen ratio}\label{nc1}\scalebox{0.8}{
		\begin{tabular}{c|ccc|ccc}
			\hline
			Methods& \multicolumn{3}{c|}{Pubmed} & \multicolumn{3}{c}{Blogcatalog} \\
			unseen-ratio         & 10\%  & 30\%  & 50\%  & 10\%  & 30\%  & 50\% \\\hline
			RawFeats       & 79.22 & 77.66 & 77.74 & \textbf{90.00} & \textbf{89.05} & \textbf{87.08}\\
			G2G            & 80.70 & 76.67 & 76.31 & 62.35 & 56.19 & 48.46\\
			GraphSAGE & 82.05 & 81.32 & 79.68 & 71.48 & 69.33 & 64.92 \\
			CADE-MS   & \textbf{84.25} & \textbf{83.40} & \textbf{81.74} & 77.35 & 73.71 & 70.88\\
			CADE-MA   & \textbf{84.56} & \textbf{83.03} & \textbf{82.40} & 84.33 & 82.21 & 79.04\\\hline
	\end{tabular}}
\end{table*}

	\begin{table}[!th]
		\centering
		\caption{Prediction results for the three datasets (micro-averaged F1 scores).}\label{nc2}\scalebox{1}{
			\begin{tabular}{ccccc}
				\hline
				& Reddit & PPI & Pubmed/30\% & Blogcatalog/30\% \\\hline
				Graph2Gauss     & 72.48  & 43.06  & 76.67 & 56.19\\
				GraphSAGE-mean  & 89.73  & 50.22 & 81.32 & 69.33\\
				CADE-MS-mean    & \textbf{92.71}  & \textbf{58.22} & \textbf{83.40}&\textbf{73.71}\\
				CADE-MA-mean    & \textbf{92.80}  & \textbf{57.17} & \textbf{83.03} & \textbf{82.21}\\
				GraphSAGE-LSTM  & 90.70  & 50.53 & \textbf{81.26}& wait\\
				CADE-MS-LSTM    & 90.48  & \textbf{56.32} & \textbf{82.37}& 70.30\\
				CADE-MA-LSTM    & \textbf{93.25}  & \textbf{54.00} & \textbf{82.65} & \textbf{82.21}\\
				GraphSAGE-pool  & 89.32  & 51.02 & 82.52 & 71.96\\
				CADE-MS-pool    & \textbf{91.59}  & \textbf{56.66} &  82.30 & \textbf{76.22}\\
				CADE-MA-pool    & \textbf{91.15}  & \textbf{57.68} & \textbf{83.40} & \textbf{85.29}\\\hline
		\end{tabular}}
	\end{table}

\subsection{Inductive node classification}
We evaluate the node classification performance of methods on the four datasets. 
On Reddit and PPI, we follow the same training/validation/testing split used in GraphSAGE. 
On Pubmed and Blogcatalog, we randomly selected 10\%/20\%/30\% nodes for training while the rest remain unseen. 
We report the averaged results over 10 random split. 

After spliting the graph dataset, the model is trained in an unsupervised manner, then the learnt model computes the embeddings for all nodes, a node classifier is trained with the embeddings of training nodes and finally the learnt classifier is evaluated with the learnt embeddings of the testing nodes, i.e the unseen nodes. 

We compare our method against 3 baselines: (i)a logistic-regression feature-based classifier (that ignores graph structure), (ii)Graph2Gauss as another unsupervised and inductive approach recenly proposed, (iii)the original hierarchical neighbor sampling and aggregating framework, GraphSAGE.

Comparation  on node classification performance on Pubmed and Blogcatalog dataset with respect to varying ratios of unseen nodes, are reported in Table \ref{nc1}.
CADE-MS and CADE-MA outperform other approaches on Pubmed. On Blogcatalog dataset, however, RawFeats performs best mainly because that, in Blogcatalog dataset, node features are not only directly extracted from a set of user-defined tags, but also are of very high dimensionality (up to 8,189). Hence extra neighborhood information is not needed.
As shown in Table \ref{nc1}, CADE-MA performs better than CADE-MS, and both outperform GraphSAGE and G2G. CADE-MA is capable of reducing high dimensionality while losing less information than CADE-MS, CADE-MA is more likely to search for the best aggregator function that can focus on those important features of nodes. As a result, the 256-dimensional embedding learnt by CADE-MA shows the cloest node classification performance to the 8k-dimensional raw features.

Comparasion among GraphSAGE, CADE and other aggregator functions  is reported in Table \ref{nc2}. Each dataset contains 30\% unseen nodes. In general, the model CADE shows significant advance to the other two state-of-art embedding learning models in node classification on four different challenging graph datasets.


\subsection{Inductive link prediction}

Link prediction task evaluates how much network structural information is preserved by embeddings. We preform the following steps: 
(1) mark some nodes as unseen from the training of embedding learning models. For Pubmed  20\% nodes are marked as unseen; 
(2) randomly hide certain percentage of edges and equal-number of non-edges as testing edge set for link prediction, and make sure not to produced any dangling node; 
(3) the rest of edges are then used to form the input graph for embedding learning and with equal number of non-edges form the training edge set for link predictor; 
(4) after training  and  inductively generation of embeddings, the training edge set and their corresponding embeddings will help to train a link predictor; 
(5) finally evaluate the performance on the testing edges by the area under the ROC curve (AUC) and the average precision (AP) scores.

Comparation on  performance with respect to varying percentage of hidden edges  are reported in Table\ref{lp}. CADE shows best link prediction performance  on both datasets.

\begin{table*}[t]
	\centering
	\caption{Link prediction results for Pubmed/PPI w.r.t different percentage of hidden-edges}\label{lp}
	\scalebox{0.8}
	{
		\begin{tabular}{cccccccccc}
			\hline
			Dataset     & Methods   & \multicolumn{2}{c}{90\%:10\%}& \multicolumn{2}{c}{80\%:20\%}& \multicolumn{2}{c}{60\%:40\%}& \multicolumn{2}{c}{40\%:60\%}\\
			&&AUC&AP&AUC&AP&AUC&AP&AUC&AP\\\hline
			\multirow{5}*{Pubmed}& RawFeats  & 57.61 & 54.72 &58.51&56.19&54.47&52.82&52.41& 50.77 \\
			~& G2G       & 64.13 & 68.60 &63.52&65.15&60.03&66.16&58.97&61.17\\
			~ & GraphSAGE & 85.49 & 82.79 &87.64&83.35& 81.07 & 77.47 & 79.34 & 74.92 \\
			~& CADE-MS& \textbf{89.95} & \textbf{88.79} &\textbf{90.36}&\textbf{86.67}& \textbf{87.14} & \textbf{83.77} & \textbf{84.76} & \textbf{79.53} \\
			~& CADE-MA& \textbf{89.73} & \textbf{89.76} &\textbf{90.94}&\textbf{88.90}& \textbf{90.54} & \textbf{87.89} & \textbf{85.27} & \textbf{80.15} \\\hline
			\multirow{4}*{PPI}& RawFeats & 57.46 & 56.99 &57.34&56.86&57.35&56.75&56.83&56.36\\
			~& G2G       & 60.62 & 58.98 & 60.99&59.38&61.05&59.54&60.93&59.49\\
			~& GraphSAGE & 82.74&81.20 & 82.21&80.66 & 82.11 & 80.51 & 82.07&80.70 \\
			~& CADE-MS & \textbf{85.87}&\textbf{85.08} & \textbf{84.21}&\textbf{83.48} & \textbf{84.46} & \textbf{82.84} & \textbf{83.61} &\textbf{82.39} \\
			~& CADE-MA & \textbf{86.33}&\textbf{85.32} & \textbf{85.85}&\textbf{84.63} & \textbf{84.15} & \textbf{82.15} & 81.98 & 79.54 \\\hline
	\end{tabular}}
\end{table*}

\subsection{Model study}\label{modelstudy}

\subsubsection{Sampling Complexity in CADE-MS}

\begin{figure*} 
	\centering 
	\subfigure{	\label{sc}
		\includegraphics[width=0.46\linewidth]{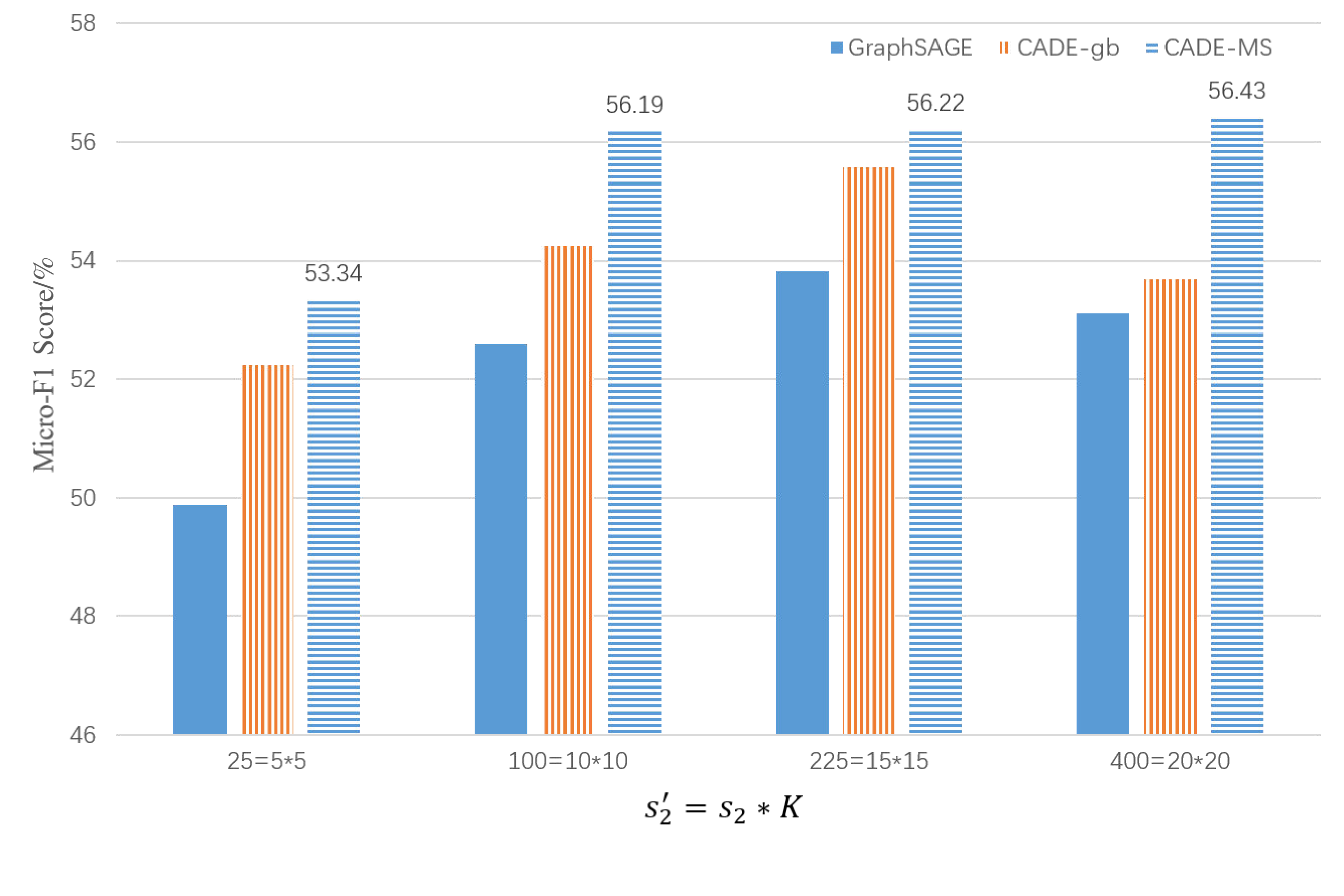}}
	\hspace{0.01\linewidth}
	\subfigure{\label{dim}
		\includegraphics[width=0.46\linewidth]{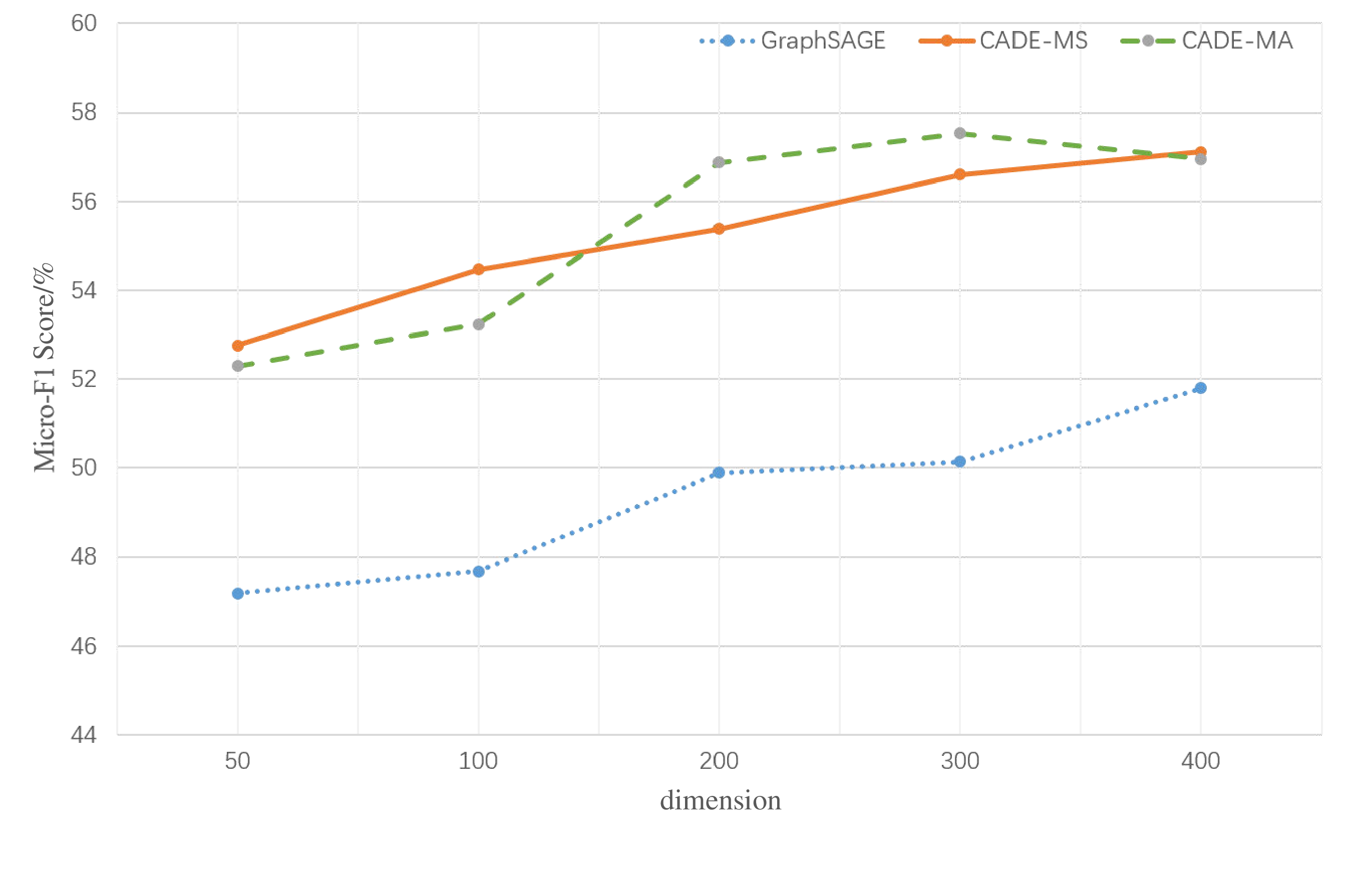}}
	\caption{(a)Node classification performance (micro-f1 score) w.r.t varying sampling sizes, (b)Classification results w.r.t different embedding dimension on PPI.}
	\label{ks}
\end{figure*}

Our proposed CADE-MS requires multiple neighborhood sampling, which increases the complexity of embedding learning. Yet by comparing CADE-MS against GraphSAGE with the same quantity of sampled neighborhood per node, the superiority of CADE model over existing models is still vast. In practice, we set the sampling layer $L=2$ and the first-layer sampling size as 20. 
For the second layer, denote by $s'_2$ the sampling size in GraphSAGE, and by $s_2$ and $T$ the sampling size and sampling time in CADE-MS. 
We compare the two methods with $s'_2=s_2*K$.

A variant of CADE, called CADE-gb, applies only memorable global bias and no dual-encoding framework, has the same sampling complexity as GraphSAGE.
For the efficiency of experiment, we conduct experiments of node classification on a small subset of PPI, denoted by \textbf{subPPI}, which includes 3 training graphs plus one validation graph and one test graph. Results are reported in Figure \ref{sc}. With much smaller sampling width,  CADE-MS still outperforms the original framework significantly. 

\ignore{
	\begin{table}[!th]
		\centering
		\caption{node classification performance (f1-micro score) w.r.t varying sampling sizes}\label{ba}
		\begin{tabular}{|c|cccc|}
			\hline
			$s'_2$      & 25    & 100    & 225   & 400  \\\hline
			GraphSAGE  & 49.87 & 52.60 & 53.82 & 53.11\\
			CADE-gb & 52.24 & 54.25 & 55.58 & 53.69\\\hline
			$s_2-K$   & 5-5   & 10-10   & 15-15 & 20-20\\\hline
			CADE-MS    & \textbf{53.34} & \textbf{56.19} & \textbf{56.22} & \textbf{56.43}\\\hline
		\end{tabular}
	\end{table}
}

It implicates that searching for the best representation match through multiple sampling and bi-attention is efficient  to filtering userful neighbor nodes without supervision from any node labels, and that the context-aware dual-encoding framework is capable of improving the inductive embedding learning ability without increasing sampling complexity.
We aslo observe that CADE-gb, the variant simply adding the memorable global bias, continually shows advance in different sampling sizes. 

\subsubsection{Sensitivity Study}

To evaluate the parameter sensitivity of CADE-MS and CADE-MA, we conduct node classification experiments on two superparameters: dimensionality and dual learning parameter K. 

For dimensionality sensitivity study, we compare GraphSAGE, CADE-MS and CADE-MA on PPI dataset, embedding vector dimension set as 50/100/200/300/400. Results are reported in Figure \ref{dim}. We observe that the node classification f1 scores of CADE-MS and CADE-MA both rise as the dimensionality increases, with a stable advance over GraphSAGE of about 5\%. Also, the result curves shows that CADE-MA is less sensitive to embedding dimensionality and achieves the max performance at 300 embedding dimensions, while CADE-MS stably performs better and better with increasing embedding size, which indicates that CADE-MS relies more on the embedding dimensionality. And again, the fact that CADE with only 50 embedding dimensions outperforms GraphSAGE with 400 embedding dimensions demonstrates the effectiveness of our proposed model.


The second sensitivity experiment is to study the influence of K, i.e. the number of candidate representation for dual encoding. We evaluate the node classification performance of CADE-MS/CADE-MA using different value of $K$. Experiments are conducted on subPPI in consideration of the huge memory cost caused by a large value of K, and we set $L=2$ and $s_1=20$, $s_2=25$. In order to only measure the influence of different $K$, we remove the global bias in CADE-MA and CADE-MS. In addition, we also report the runtime of the model. The results are shown in Figure \ref{ks}. 

\begin{figure*} 
	\centering 
	\subfigure[Reddit]{\label{ks-ms}
		\includegraphics[width=0.46\linewidth]{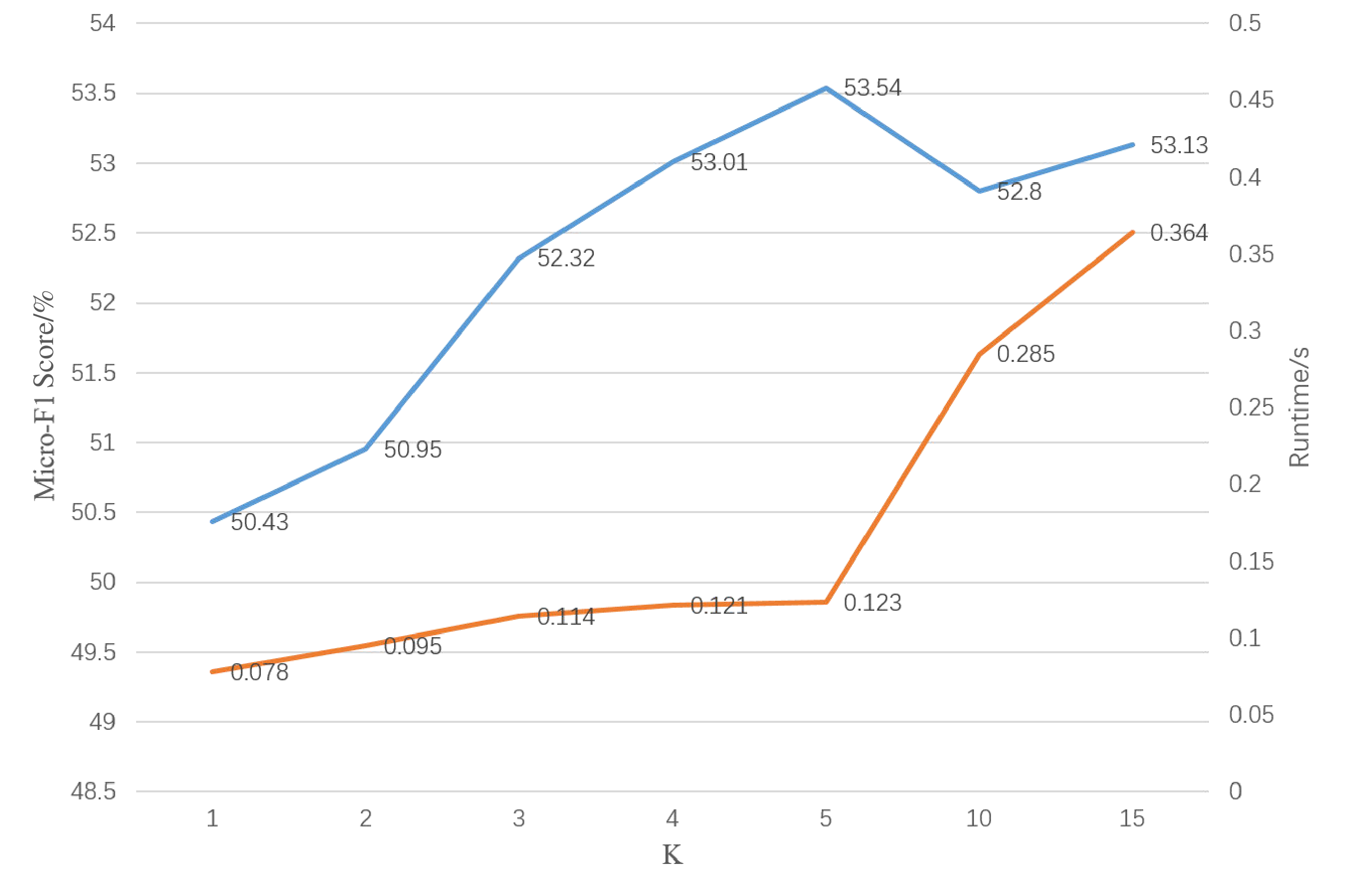}}
	\hspace{0.01\linewidth}
	\subfigure[PPI]{\label{ks-ma}
		\includegraphics[width=0.46\linewidth]{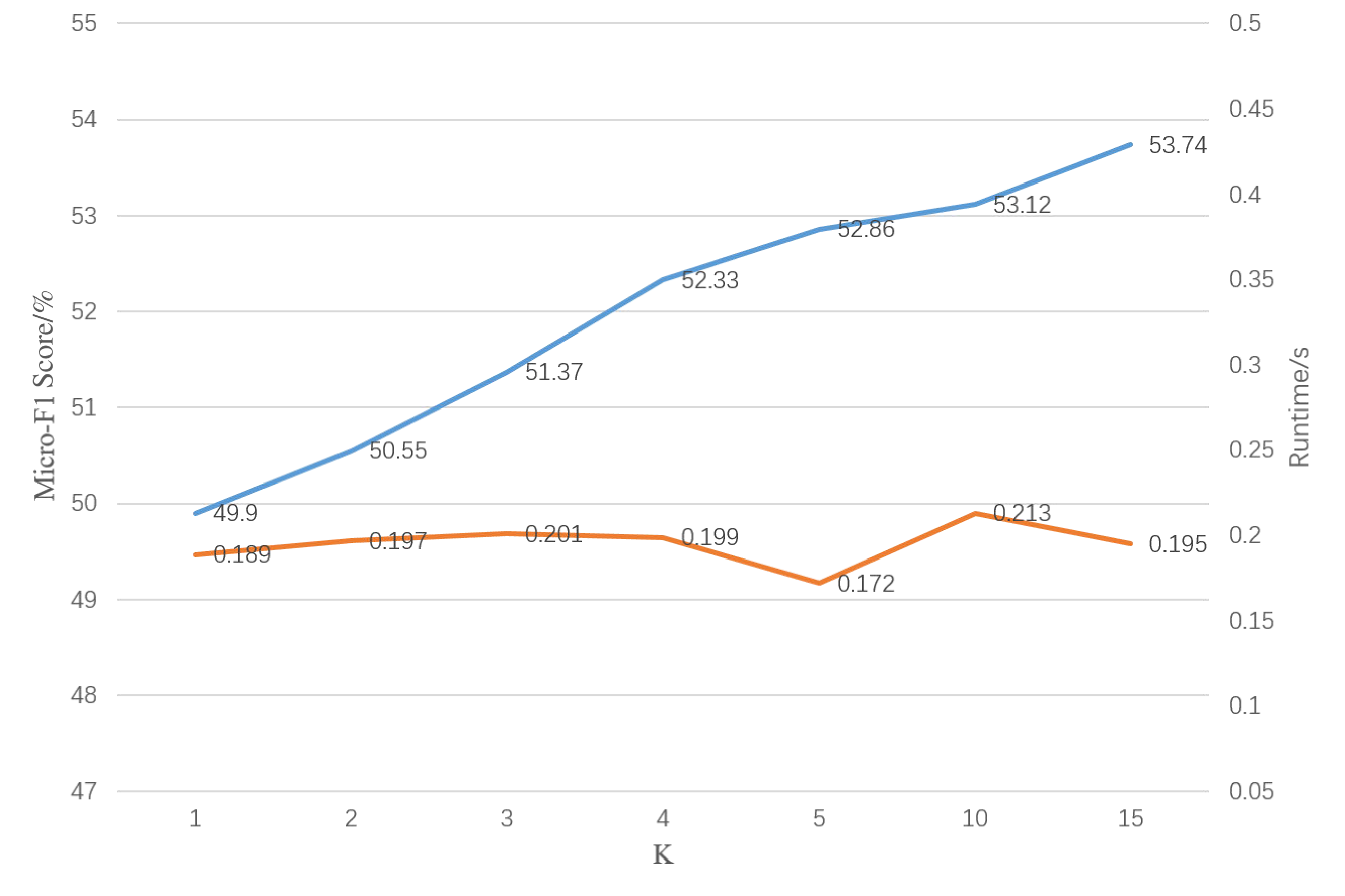}}
	\caption{Classification results (micro-averaged F1 scores) w.r.t different values of K on subPPI.}
	\label{ks}
\end{figure*}

As we can see from Figure \ref{ks-ms}, the runtime of CADE-MS increases as K increases, while at K=5 the F1 scores already reach the highest value, which indicates that setting K as 5 is the most effective and efficient. In the implementation of CADE-MA, we calculate K aggregator functions parallelly, which is why we can see in Figure \ref{ks-ma} that the runtime of CADE-MA is independent of K, while the node classification performance of CADE-MS keeps improving as K increases.

%
%
%
%

\section{CONCLUSION}
We proposed CADE, an unsupervised and inductive network embedding approach which learned and memorized global identities for seen nodes and was generalized to unseen nodes. We applied a bi-attention architeture upon hierarchical aggregating layers to  capture the most relevant representations dually for any positive pair. We effectively combined inductive and transductive ideas by allowing trainable global embedding bias to be retrieved in hidden layers. Experiments demonstrated the superiority of CADE over baselines on unsupervised and inductive tasks. In the future, we would use dual encoding framework in supervised embedding learning, or combing dual encoding with G2G by learning distribution representations dually for positive pairs. It would be also interesting to employ our approach in symbolic searching related fields, such as planning with incomplete domains (c.f., \cite{DBLP:journals/ai/ZhuoK17}), logic based domain model learning (c.f., \cite{DBLP:journals/ai/ZhuoM014,DBLP:journals/ai/Zhuo014}) and shallow model-based plan recognition (c.f., \cite{DBLP:journals/tist/Zhuo20,DBLP:conf/atal/TianZK16}) based on graph embeddings of propositions and actions.

\acks{We  thank the support of the National Key R$\&$D Program of China (No. 2018YFB0204300), the National Natural Science Foundation of China  (No. 11701592 for Young Scientists of China, No. U1811263 for the Joint Funds and No. U1611262),  the Guangdong Natural Science Funds for Distinguished Young Scholar (No. 2017A030306028), the Guangdong special branch plans young talent with scientific and technological innovation and the Pearl River Science and Technology New Star of Guangzhou.}

\bibliography{DGRL}

%
%
%
%

\end{document}

%% file: math_commands.tex
\usepackage{amsmath,amsfonts,bm}

\def\eqref#1{equation~\ref{#1}}

\def\1{\bm{1}}

\def\vb{{\bm{b}}}

\def\vh{{\bm{h}}}

\def\vz{{\bm{z}}}

\def\mB{{\bm{B}}}

\def\mW{{\bm{W}}}
\def\mX{{\bm{X}}}

\DeclareMathAlphabet{\mathsfit}{\encodingdefault}{\sfdefault}{m}{sl}
\SetMathAlphabet{\mathsfit}{bold}{\encodingdefault}{\sfdefault}{bx}{n}

\def\gG{{\mathcal{G}}}

\def\gS{{\mathcal{S}}}

\def\sN{{\mathbb{N}}}

\def\sP{{\mathbb{P}}}

\def\sZ{{\mathbb{Z}}}

\newcommand{\E}{\mathbb{E}}

\newcommand{\R}{\mathbb{R}}

